# Analysis of Drug repurposing Knowledge graphs for Covid-19

Ajay Kumar Gogineni [†]
ECE
Virginia Tech
gogineniak20@vt.edu

## ABSTRACT

Knowledge graph (KG) is used to represent data in terms of entities and structural relations between the entities. This representation can be used to solve complex problems such as recommendation systems and question answering. In this study, a set of candidate drugs for COVID-19 are proposed by using Drug repurposing knowledge graph (DRKG). DRKG is a biological knowledge graph constructed using a vast amount of open source biomedical knowledge to understand the mechanism of compounds and the related biological functions. Node and relation embeddings are learned using knowledge graph embedding models and neural network and attention related models. Different models are used to get the node embedding by changing the objective of the model. These embeddings are later used to predict if a candidate drug is effective to treat a disease or how likely it is for a drug to bind to a protein associated to a disease which can be modelled as a link prediction task between two nodes. RESCAL performed the best on the test dataset in terms of MR, MRR and Hits@3.

## KEYWORDS

Knowledge Graph, neural network, attention

## 1  Introduction

In the initial stages of a viral outbreak such as Covid 19, a vaccine or therapies that are effective in controlling the outbreak will typically not be available. To reduce the impact, finding a vaccine will have the most importance. An important question to answer in this context is which compounds should be selected and tested to be a plausible remedy. Scientists and doctors analyse the gene structure of the virus and shortlist some compounds based on their experience and initial tests. These compounds will then be tested rigorously by FDA to check if the compound is safe and is a good drug that can inhibit the gene of the virus. Developing a novel drug is time consuming and costly.

Since a vaccine or a suitable therapy is not available for treating Covid-19 at the initial stages, drugs approved for other diseases were repurposed such that they can be used for Covid-19. Some of these drugs are clinically approved by the FDA and a lot of information about the drugs is already available.

In comparison to building a drug from scratch, it requires significantly less time and capital [1]. Among the drugs analysed, Remdesevir [2] was approved by FDA. Use of biological knowledge graphs for drug repurposing has been emphasized in the articles [3].
Instead of depending on previous experience and intuition, a knowledge graph can be constructed and analysed to find drugs that are effective on a virus.

## 2  Background

Drug repurposing is typically formulated as a knowledge completion task where we try to predict all the missing edges in a graph [4]. Knowledge graph is a collection of head, relation and tail (h, r, t) triples. The head and tail nodes are related to each other by the relation, r. Knowledge graph completion can be formulated as link prediction where the model tries to predict the existence of a relation between head and tail nodes. One more way to formulate it is to infer the tail node when the head node a particular relation is specified. It can also be used to detect head nodes when tail nodes and relations are given.

Recent research has focused on building knowledge graph (KG) embedding methods for knowledge graph completion, which map the nodes and edges to embedding space while preserving the structural information of the KG and can be modelled using machine learning techniques. A survey on the available models can be found in this paper [5]. Embeddings are learned while minimizing an objective function, which is called a scoring function. Scoring functions can be categorized into distance based and semantic similarity based functions as described in [6]. TransE [7], TransH and RotatE [8] fall under distance based function. RESCAL [9] , DistMult [10], ComplEx[11], and HolE fall under semantic similarity based scoring function. Training time and inference time and model complexity are different since some scoring functions are computationally expensive.

Few interactions between proteins and candidate compounds that can inhibit the gene of Covid-19 are present in the DRKG dataset. The authors in [12] modelled a few shot link predictions where they predict candidate drugs that can inhibit the target gene.

The figure below shows an example of how knowledge can be represented as a graph. The edges between the nodes show how one node is related to another and the edges are sensitive to direction.

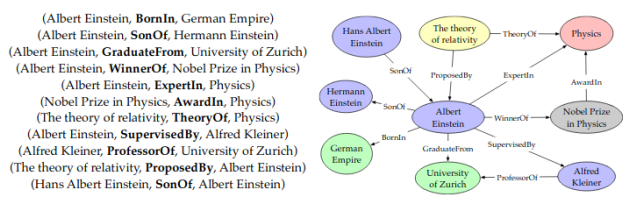

Fig. 1: Demonstration of how a knowledge graph can be constructed from a knowledge base. Source : [https://ryanong.co.uk/2020/11/06/day-311-ryans-phd-journey-overview-of-knowledge-graphs/]

## 3 Data

DRKG [13] is a combination of the following medical datasets: DrugBank, Hetionet, GNBR, String, IntAct and DGIdb, and data related to Covid19 from related publications. The dataset contains 97,238 entities which belong to 13 categories; and 5,874,261 triplets. There are 107 edge-types in this dataset.

The data is represented in the form of (h, r, t) triplets, where h is the head node, t is tail node and r is the relation. The figure below shows a sample of DRKG and gives an idea on how nodes in the KG interact. The nodes contain genes, compounds, diseases, side effects of compounds.

The dataset contains the following information :
All the triplets in a tsv file, a dictionary that maps entities to data sources, and dictionaries that map entities and relations to numbers.

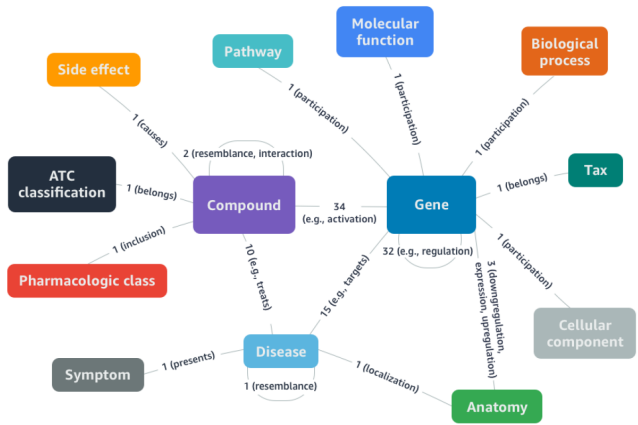

Fig. 2: Figure representing how entities interact in DRKG. The number on the edges denote how many types of relation exist between the corresponding nodes. Source : [https://www.dgl.ai/news/2020/06/09/covid.html]

## 4 Previous Work

Previous work used TransE to learn embeddings of the entities and the relationships. h, r and t represent the embeddings of head, relation and tail entities respectively. In the TransE model is that, for a plausible triple, we want the summation of head and relation embedding to be close to the tail node embedding for a valid triplet. The figure below shows how the score is calculated for the TransE model.

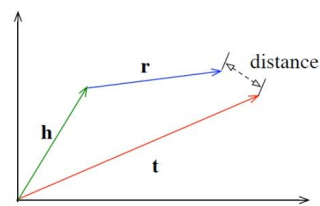

Fig. 3 TransE score calculation.

The process of representing entities and relations as embedding vectors is called Knowledge Graph Embedding (KGE). The current work aims to tackle drug repurposing and validate performance of various KGE models similar to TransE. The embeddings are regularized by L2 norm. Another variant of the TransE model uses L1 normalization. This model is called TransE L1.

A comprehensive review of such models can be found in this paper [12], where the authors tune the following four parameters : 1) Embedding space, 2) Scoring function, 3)

Encoding models : Generate node embeddings using neural networks and 4) Using Auxiliary information (incorporating textual and visual descriptors with the KG to add more context).

The embedding space is selected such that it is differentiable and should be able to define a scoring function. The score of a correct (h, r, t) triplet is high.

# 5 Models

Models based on Scoring Function:

5.1 RotatE : It models r vectors as a rotation in a complex plane.

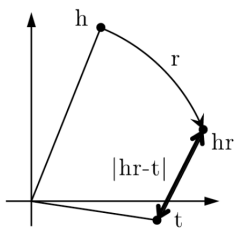

Source : [https://aws-dglke.readthedocs.io/en/latest/kg.html]

The magnitude of r vector is one. Hence, r vector can be represented as $e^{i\theta}$. The r vector rotates h vector by an angle θ in a counter clockwise direction. The score is represented by the following equation, where the circle represents element wise multiplication :

$$d_r(h, t) = \|h \circ r - t\|$$

5.2 RESCAL

It is a bilinear model and maps a relation in the form of a symmetric matrix to model pairwise interactions between entities. The model has a large number of parameters and hence it might overfit on the training dataset. And the model is computationally expensive and can only model symmetric relations. $M_r$ is the relationship matrix, d is the dimension of node embedding vectors. h is the embedding vector for head node and t is the embedding vector for the tail node. The following equation represents the scoring function for RESCAL.

$$f_r(h,t) = \mathbf{h}^\top M_r t = \sum_{i=0}^{d-1}\sum_{j=0}^{d-1}[M_r]_{ij}.[h]_i.[t]_j$$

5.3 DistMult :

To reduce the number of parameters in the RESCAL model, Distmult was proposed which modifies the scoring function. It only uses diagonal elements in the matrix and hence has fewer parameters. The drawback of this model is that the triples : (h, r, t) and (t, r, h) will be given the same score since the score can be considered as a dot product between three vectors. The following equation represents the scoring function for DistMult.

$$f_r(h,t) = \mathbf{h}^\top diag(r) t = \sum_{i=0}^{d-1}[r]_i.[h]_i.[t]_i$$

5.4 ComplEx

This model can capture antisymmetric relations and is not computationally expensive. To model asymmetric relations, this model uses complex space to encode nodes and relations. The matrix is symmetric, but the embedding of a node when considered as head and tail are complex conjugates of one another.

5.5 CompGCN with ConvE scoring function.

CompGCN is a graph convolutional framework which was proposed in [**14**] and learns representations for nodes and relations at the same time. In CompGCN, the embedding of a central node is modified by applying a composition operation over the surrounding nodes. This helps the model to get more contextual information while embedding the nodes. The size of the model increases in a linear w.r.t number of relations. The code to train the model is available in this github repository : [https://github.com/uma-pi1/kge]

When ConvE is used as a scoring function in CompGCN model, the embeddings are initialized and then converted into 2d matrices which can be thought of as images as proposed in [**15**]. Convolutional filters are applied on these images to learn features and make predictions. This model extends the ideas from convolutional neural networks to KG embedding models.

The figure below, taken from [https://arxiv.org/abs/1707.01476] demonstrates how CompGCN model can be applied simultaneously to head and relation embedding. The final layer of the model predicts the tail entity for the corresponding head and relation embedding.

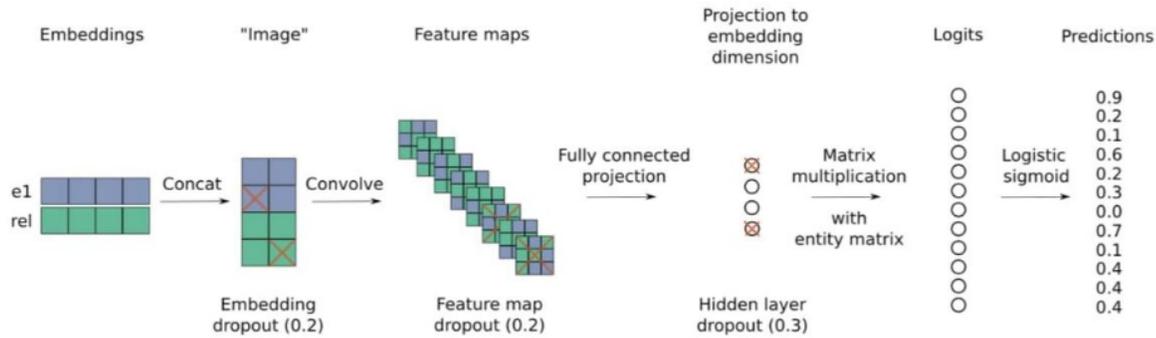

Fig 3. The head and relation embeddings are concatenated as given as input to a convolutional neural network. Source : Convolutional 2D Knowledge Graph Embeddings [https://arxiv.org/abs/1707.01476]

## 5.6 Transformer model

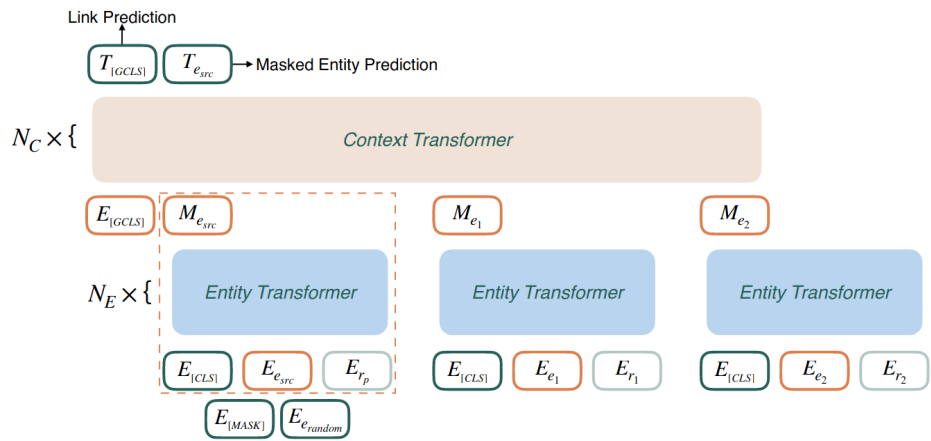

Fig 4. Transformer architecture Source : Hitter [https://arxiv.org/pdf/2008.12813.pdf]

Similar to CompGCN, the transformer model proposed in [16] learns the node and relation embedding jointly. Self attention in the transformer model is similar to the composition operation in GCN where information from context is aggregated to represent the current node. The model has two transformer blocks: Given a head node, the first block extracts features corresponding to the head node and the relation associated with it. And the second block aggregates the information obtained from the first block and the neighboring blocks. All the blocks with a connection to the head node are considered as the context or the neighboring blocks.

The transformer model captures information from the context of a node very effectively when compared to GCN models. The main drawback of the model is that it gets the representation using only the head node and relation in the first step. The model then calculates the dot product of this representation with all target entities. The inference time for the model is less.

Figure 4. shows the model architecture proposed in [16]. The two blocks are arranged hierarchically. The first block is called entity transformer and takes a (head, relation) pair as input. The second block is called the Context transformer and it aggregates the information obtained from the neighbouring nodes.

Link prediction can be directly done without using information from the context of the node. The embedding vector, $M_{e_{src}}$ can be directly used to predict the tail node given head node and the relation. This model can be used as a baseline since the number of parameters are very less compared to the original model with two blocks. The model is trained by completing the following two tasks:

1. For a given node and relation embeddings, the context transformer returns $T_{[GCLS]}$, an embedding vector that is used to predict the tail node.
2. The second task is similar to masked language modeling employed in BERT [17]. This is called Masked Entity prediction task (MEP). In this task, some of the head node embeddings are randomly replaced by a special token called $E_{[MASK]}$ or $E_{e_{random}}$. In this scenario, $T_{e_{src}}$ embedding vector is used to detect the head node that was changed.

The model should not depend entirely on the context or discard the information obtained from the context nodes completely. A MEP task is very crucial for the model to give equal importance to context and the input embeddings, since it forces the model to not depend entirely on the context. When some of the nodes are masked, the model learns to predict the masked node by using the contextual information.

Some nodes can have connections with a large fraction of the dataset. To avoid this, nodes are sampled such that only a portion of the neighboring nodes will show up in the training data. Once the embedding vector, $M_{e_{src}}$ is obtained for a given head node and relation, its dot product is taken with all the candidate tail nodes. This dot product is similar to the score of a given triplet. These scores are normalized using softmax activation function and hence a probability distribution over all the possible tail nodes is obtained. The model is trained by minimizing the cross entropy loss obtained from the above distribution.

Adam optimizer is used to train the model with early stopping and learning rate warmups as callbacks. The encoder block has three layers, i.e. $N_E$ = 3 in the transformer architecture where each encoder uses multi head attention with 8 heads. The code to train the model and the required input format is available in this repository [18]:

## 6 Quality of Embedding : Metrics

Once the node and relation embeddings are generated, the quality of the embeddings can be assessed using the following metrics : Mean Rank (MR), Mean Reciprocal Rank (MRR), Hits at 1 (H@1), Hits at 3 (H@3), and Hits at 10 (H@10). These metrics can be calculated by following the three steps given below:

1. For each possible triple (hi,ri,ti) intest set, remove the head node, hi.

2. Compute the score f (h,ri,ti), for every possible head in the entire data set.

3. Sort the scores obtained in Step2 in decreasing order and store the index of the correct head node which is the rank of the corresponding head node, hi.

$$MR = \frac{1}{n}\sum_{i=1}^{n} \text{rank}_i, \quad MRR = \frac{1}{n}\sum_{i=1}^{n} \frac{1}{\text{rank}_i}, \quad H@N = \frac{1}{n}\sum_{i=1}^{n} \mathbf{1}\{\text{rank}_i \leq N\}$$

Source : [http://cs229.stanford.edu/proj2019aut/data/assignment_308832_raw/26646712.pdf]

Figure Representing mathematical equations of metrics. H@N indicated how many times the correct head node is in the top N predictions of the model. MR represents the average index at which the correct head node is predicted by the model. MRR is the harmonic mean of the rank.

## 7 Training

DGL-KE [18], an open source library for KGE is used in this project to get node and relation embeddings. All the valid triples form positive examples in the dataset. To get negative examples, some head and tail nodes are replaced by randomly selected head and tail nodes from the dataset respectively. The models are trained to return a high score for a correct triplet and a low score for an incorrect triplet.

Logistic loss is used to train the models where the target, y is +1 for positive examples and -1 for negative examples. f (h, r, t) is the score for the given triple and $D^+$ and $D^-$ are the negative and positive examples respectively

Optimization equation for logistic loss :

$$\text{minimize} \sum_{(h,r,t) \in \mathbb{D}^+ \cup \mathbb{D}^-} \log(1 + e^{-y \times f(h,r,t)})$$

Source : [https://aws-dglke.readthedocs.io/en/latest/kg.html]

## 8   Results

All the models use an embedding dimension of 400 except the ConvE model which uses an embedding dimension of 50.

The models will predict the best candidate drug that has a treatment relationship with disease entities related to Covid-19. 8104 FDA approved drugs with molecular weight less than 250 from DRKG dataset are used as head nodes. 34 Covid-19 related proteins in DRKG are selected as target nodes. The treatment is modelled as a relation between head node which is candidate drug and tail node which is the proteins related to Covid-19.

All possible (head node, relation, tail node) triplets are scored and sorted in ascending order. The top 100 drug candidates based on score are considered as the predictions of a particular model.

Some of the models such as CompGCN and RESCAL have a large number of parameters and take a lot of time to train.

The drug that is common among the model predictions will be a strong drug candidate. The model predictions are validated against the drugs that went to clinical trials in the initial stages of the outbreak.

| Model | Mean Rank | Mean Reciprocal Rank | Hits @ N =3 |
|---|---|---|---|
| TransE | 66.3 | 0.407 | 0.477 |
| RotateE | 100 | 0.45 | 0.5 |
| RESCAL | ***57*** | ***0.66*** | ***0.7*** |
| DistMult | 76.8 | 0.48 | 0.52 |
| ComplEx | 83 | 0.65 | 0.7 |

Table1. Showing metrics on test dataset for models based on scoring function. RESCAL outperforms all the models across all three metrics.

## 9   Drug Predictions

| Scoring function | Drugs Proposed |
|---|---|
| TransE L2 | **Ribavirin**<br>Methylprednisolone<br>**Colchicine** |
| TransE L1 | **Ribavirin**<br>Methylprednisolone<br>**Colchicine**<br>**Chloroquine** |
| RotateE | **Colchicine**<br>Deferoxamine<br>Mavrilimumab |
| RESCAL | **Colchicine**<br>Deferoxamine<br>Mavrilimumab |
| DistMult | Mavrilimumab<br>Bevacizumab<br>Thalidomide |
| ComplEx | **Ribavirin**<br>**Chloroquine**<br>Tetrandrine |

| Model based on encoder | Drugs Proposed |
|---|---|
| CompGCN with ConvE scoring function | Azithromycin<br>Ribavirin |
| Transformer | Dexamethasone<br>**Chloroquine**<br>**Hydroxychloroquine**<br>Bevacizumab |

Table 2. The table above represents the drug candidates proposed by the models analysed in this study. The drugs that are represented in bold showed positive results in clinical trials and are also proposed by other drug repurposing articles.

## 10   Summary

Overall, developing a novel KGE model is to find the best KGE model based on embedding space, scoring function and models to encode the information for drug repurposing. RESCAL has the largest number of parameters and performs best on the test dataset in terms of evaluation metrics. Transformer model proposed Chloroquine and Hydroxychloroquine as the candidate drugs for Covid-19. Large training data is not always available and hence models that perform better on smaller datasets and fewer parameters should be given more importance in future work. This analysis shows the importance and usefulness of node and relation embeddings on DRKG dataset. Hyperparameter tuning and selecting the appropriate model is crucial for this study.